\documentclass[letterpaper]{article} 
\usepackage{aaai25}  
\usepackage{times}  
\usepackage{helvet}  
\usepackage{courier}  
\usepackage[hyphens]{url}  
\usepackage{graphicx} 
\usepackage{natbib}  
\usepackage{caption} 
\usepackage{algorithm}
\usepackage{algorithmic}
\usepackage{multirow}
\usepackage{amsmath}
\usepackage{tabularx}
\usepackage{xcolor}
\usepackage{xltabular}
\usepackage{ltcaption}
\usepackage{longtable}
\usepackage{subcaption}
\usepackage{appendix}

\usepackage{newfloat}
\usepackage{listings}
\DeclareCaptionStyle{ruled}{labelfont=normalfont,labelsep=colon,strut=off} 
\floatstyle{ruled}
\newfloat{listing}{tb}{lst}{}
\floatname{listing}{Listing}
\title{CareBot: A Pioneering Full-Process Open-Source Medical Language Model}
\author{
    Lulu Zhao\textsuperscript{\rm 1}\thanks{Corresponding author},
    Weihao Zeng\textsuperscript{\rm 2},
    Xiaofeng Shi\textsuperscript{\rm 1},
    Hua Zhou\textsuperscript{\rm 1}
}
\affiliations{
    \textsuperscript{\rm 1}Beijing Academy of Artificial Intelligence (BAAI)\\
    \textsuperscript{\rm 2}School of Artificial Intelligence, Beijing University of Posts and Telecommunications \\
    \texttt{llzhao@baai.ac.cn} \\

}

\usepackage{bibentry}

\begin{document}

\maketitle

\begin{abstract}
Recently, both closed-source and open-source LLMs have made significant strides, outperforming humans in various general domains. However, their performance in specific professional domains such as medicine, especially within the open-source community, remains suboptimal due to the complexity of medical knowledge. In this paper, we propose CareBot, a bilingual medical LLM, which leverages a comprehensive approach integrating continuous pre-training (CPT), supervised fine-tuning (SFT), and reinforcement learning with human feedback (RLHF). Our novel two-stage CPT method, comprising Stable CPT and Boost CPT, effectively bridges the gap between general and domain-specific data, facilitating a smooth transition from pre-training to fine-tuning and enhancing domain knowledge progressively. We also introduce DataRater, a model designed to assess data quality during CPT, ensuring that the training data is both accurate and relevant. For SFT, we develope a large and diverse bilingual dataset, along with ConFilter, a metric to enhance multi-turn dialogue quality, which is crucial to improving the model's ability to handle more complex dialogues. The combination of high-quality data sources and innovative techniques significantly improves CareBot's performance across a range of medical applications. Our rigorous evaluations on Chinese and English benchmarks confirm CareBot's effectiveness in medical consultation and education. These advancements not only address current limitations in medical LLMs but also set a new standard for developing effective and reliable open-source models in the medical domain\footnote{https://github.com/FlagOpen/CareBot}. 
\end{abstract}

%

\section{Introduction}
Recently, the advent of generative large language models (LLMs) like ChatGPT \cite{ChatGPT} and LLaMA \cite{LLaMA1,LLaMA2} has revolutionized human-computer interaction. These models excel at basic text understanding and complex problem-solving tasks, demonstrating capabilities akin to human understanding and reasoning. However, in industrial applications, the professionalism and cost-effectiveness of LLMs are more concerned.  Although a series of closed-source models such as GPT-4 still perform well in specialized domains, considering the risk of data privacy, it is not convenient to use such APIs to handle domain-specific issues. In the open-source community, a lack of domain-specific knowledge often limits the performance of open-source models in specialized areas, such as medical \cite{Baichuan2,DoctorGLM,BioMistral}. The complexity and depth of medical knowledge present significant challenges for developing accurate and secure medical LLMs. Nonetheless, we believe that medical LLMs hold immense potential and can significantly contribute to diagnostic assistance, consultation, drug recommendation, and more. Thus, developing a fully open-source LLM tailored for the medical domain is of paramount importance.

Currently, there are several medical LLMs available in this domain . However, most of these models rely solely SFT \cite{zhang2023huatuogpttaminglanguagemodel,zhang2024ultramedicalbuildingspecializedgeneralists,han2023medalpacaopensourcecollection}. As is well-known, pre-training is a critical phase for learning domain-specific knowledge, and depending exclusively on SFT results in models that can only produce answers in a fixed format. Another approach attempts to integrate pre-training with SFT by converting pre-training data in specific domains into a unified format similar to SFT data, such as (instruction, output) pairs using GPT-3.5 \cite{chen2023huatuogptiionestagetrainingmedical}. This method of synthesizing large amounts of data can lead to the inclusion of significant amounts of incorrect knowledge that aligns with GPT-4 but diverges from human expertise, as well as high data synthesis costs. Furthermore, \citet{yang2023zhongjingenhancingchinesemedical} introduce Zhongjing for Chinese medicine, which employs to implement the pipeline training from pre-training, SFT, to RLHF. However, this approach involves two phases of transformation for the base model, which may lead to issues such as catastrophic forgetting or model degradation \cite{cheng2024adapting}. Additionally, previous efforts have predominantly focused on data construction during the SFT stage \cite{li2023huatuo26mlargescalechinesemedical,zhang2024alpacareinstructiontunedlargelanguagemodels}, while neglecting the importance of data construction during the CPT stage. Yet, a well-designed CPT data strategy is also crucial for inserting medical expertise into the model.

To address these challenges, we propose \textbf{CareBot}, a bilingual medical LLM based on LLaMA3-8B, designed to effectively assist doctors with diagnosis, provide personalized treatment plans, and support medical education. Our approach also implements the entire process from CPT, SFT to RLHF. Most importantly, we develope a novel two-stage CPT method, consisting of stable CPT and boost CPT. The stable CPT addresses the distribution discrepancy between general and domain-specific data, while boost CPT narrows the gap between pre-training and fine-tuning data. This method facilitates a smooth transition for the model from general data to domain-specific data, and finally to fine-tuning data, thereby enhancing its domain knowledge progressively. Recognizing that data quality is critical to model performance, we also design a data quality assessment model for CPT called \textbf{DataRater}. This model employs a comprehensive quality assessment standard, evaluating aspects such as grammatical accuracy, information density, semantic consistency, and domain-specific attributes. DataRater effectively mitigates data bias, ensuring CareBot's performance and generalization capabilities in the medical domain. For SFT stage, we further construct a highly diverse medical SFT dataset, comprising single-turn and multi-turn medical dialogues, as well as medical subject knowledge multiple-choice questions, covering over 15+ departments and 100+ disease specialties. Noted that this corpus is the largest open-source bilingual medical SFT dataset available. It supports a variety of medical applications, clinical tasks, and online consultations, significantly enhancing CareBot's performance across multiple dimensions. Given the importance of data quality during the SFT stage\cite{zhou2023lima}, we employ various selection methods. One key innovation is \textbf{ConFilter}, a metric designed to measure the correlation between multiple turns, which helps in filtering multi-turn dialogues. The inclusion of high-quality multi-turn dialogues not only improves the model's ability to understand user intent and generate relevant responses but also ensures a natural, smooth dialogue experience that enhances user comfort and satisfaction. Our SFT data is sourced partly from real-world medical diagnosis dialogues and partly from GPT-3.5 generated content. This combination ensures that the model delivers informative, clear, and logically consistent answers, while also providing professional and personalized consultations akin to those of medical experts. Finally, in the RLHF stage, we leverage GPT-4 to create positive and negative medical data pairs based on the SFT results. We then apply the Direct Preference Optimization (DPO) \cite{rafailov2023direct} algorithm to align the model's output with human expression styles, providing personalized answers and recommendations, and enhancing overall user experience and satisfaction.

\begin{figure*}[t]
\centering
\includegraphics[width=14cm, height=6.5cm]{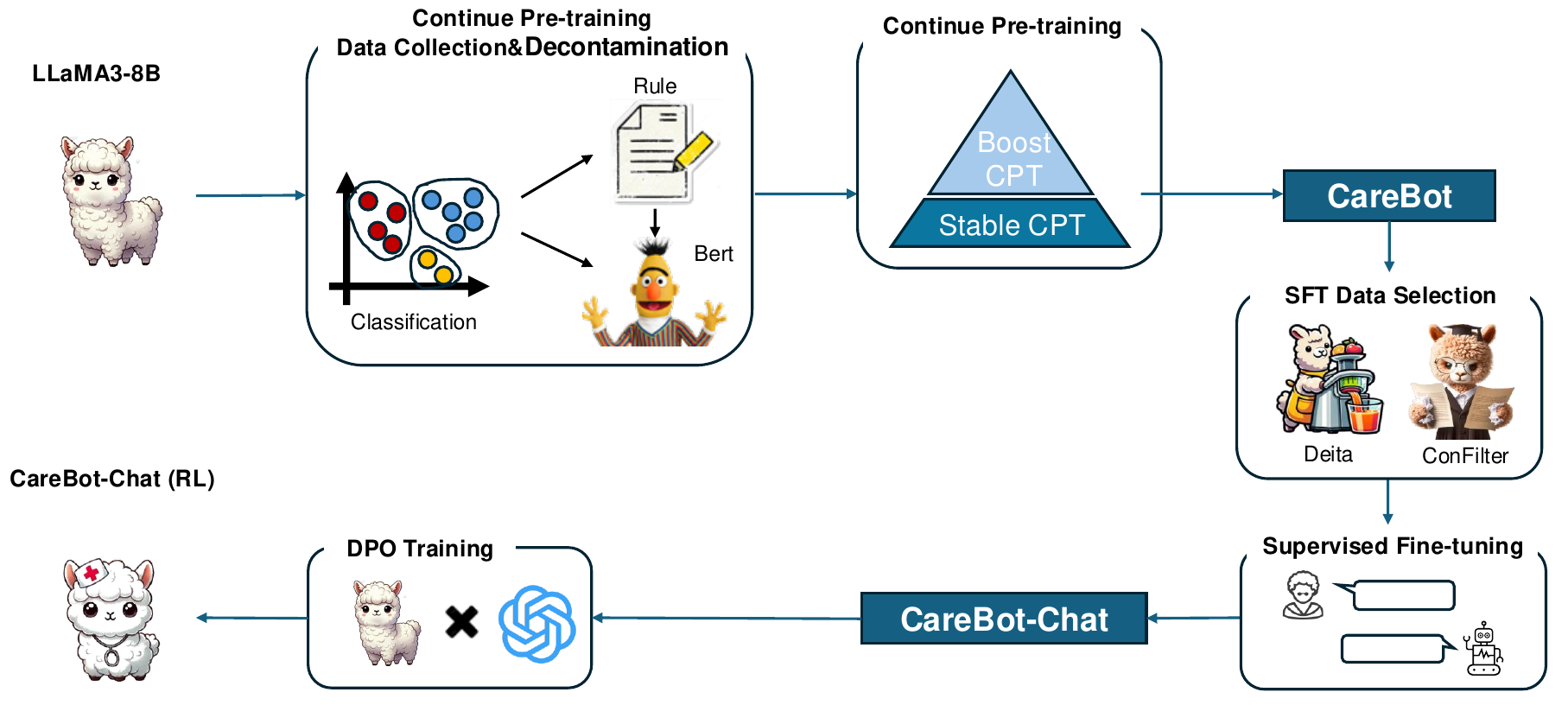}

\caption{The overall pipline of CareBot-Chat (RL), which includes the two-stage continue pre-training, supervised fine-tuning, and the DPO process.}
\label{fig:pipline}
\end{figure*} 
After extensive training and optimization, we successfully develope the CareBot. We rigorously evaluate the performance of our model using widely used Chinese and English benchmarks in medical domain. The experimental results demonstrate that CareBot excels in both medical consultation and teaching, validating that our constructed datasets significantly enhance the model's performance across multiple dimensions. The main contributions of this paper are as follows: (1) We design a novel two-stage CPT strategy that progressively and stably integrates domain knowledge into the LLM, and effectively addresses data bias and language imbalances in the original pre-training data. (2) We propose DataRater, a model for assessing data quality during CPT, ensuring that the data used for CPT  is of high quality. (3) We construct a comprehensive open-source medical SFT dataset with high data diversity and data quality. Additionally, we develope ConFilter, a metric for measuring the correlation between multiple turns, to ensure the quality of multi-turn dialogues. (4) We conduct experiments across multiple Chinese and English benchmarks to validate the effectiveness and reliability of our training strategy and datasets.

 \subsection{Continue Pre-training}
\subsubsection{Data Collection and Decontamination}

To optimize the use of existing general data resources and minimize the cost of acquiring new medical-related data, we aim to extract medical-specific pre-training data from 15T widely-covered general corpus. These datasets include web content, encyclopedias, books, and academic papers, such as C4 \cite{raffel2023exploringlimitstransferlearning}, Pile, Wudao, and PubMed, etc. To ensure the high quality of the domain-specific data, we implement a rigorous collection process that includes domain classification and quality assessment\footnote{It is worth noting that our data processing methodology is also applicable to any domain and it will be made publicly available.}.

\textbf{Domain Classification}\quad Since the general pre-training corpus is sourced from diverse datasets and lacks clear domain labels, we first conduct domain classification to extract high-quality medical data. Specifically, we sample 40k data from general corpus and use GPT-4 to perform two rounds of domain labeling to enhance accuracy. Data with inconsistent labels across two rounds is removed, leaving us with 36k high-quality seed data. We observe that certain categories, such as artificial intelligence and computers, have long tails. To address this imbalance, we utilize GPT-4 to generate additional synthetic data for these long-tail categories. Finally, we design a domain classifier based on the bge-m3\footnote{https://huggingface.co/BAAI/bge-m3}. We try a variety of multilingual pre-training models to train domain classifiers, details can be found in the Appendix A.

\textbf{Rule-based Data Quality Filtering}\quad Quality filtering is a crucial step in processing pre-training corpus. To eliminate noisy data, we use a rule-based filtering solution, including rules for removing data with insufficient tokens, excessive special characters, toxic content, and private information.

\textbf{LLM-based Data Quality Filtering}\quad By sampling and evaluating the data after rule filtering, we identify several issues: (1) The data includes advertising and marketing content, which could significantly skew the model's output preferences; (2) The data contains grammatical errors, semantic inconsistencies, and spliced, unrelated content, as well as image and video clips. Such data is detrimental to model training as it provides minimal valuable information for autoregressive learning. To address these issues, we design a data quality assessment model, DataRater, to assess data quality in terms of grammatical accuracy, information density, semantic consistency, and domain relevance, and further filter out low-quality content. Specifically, we extract 20k data from the rule-based filtered data and score them twice using the GPT-4, ranging from 0 to 5. Data with a score discrepancy of about 2 points between the two assessments is removed, resulting in a final set of 15k high-quality training data. Finally, we train the DataRater based on the bge-m3. We also try a variety of multilingual pre-training models, details can be found in the Appendix B.

\subsubsection{Training Strategy}
To gradually align the data distribution between pre-training and fine-tuning and minimize the loss of knowledge acquired during pre-training, we design a novel two-stage CPT strategy. This approach ensures a stable integration of medical knowledge into the LLM.

\noindent\textbf{Stable CPT}\quad To balance medical domain knowledge with general knowledge, we first implement a Stable CPT stage, which ensures the model maintains and enhances its general language understanding while concentrating on medical information. In this stage, we combine a high-quality medical pre-training corpus with general data via the ratio as 19:1, with a token-level distribution of 1:9 for Chinese:English. We conduct adequate experiments to search both ratios, detailed results are available in the Appendix C.

\noindent\textbf{Boost CPT}\quad To integrate medical knowledge during the model pre-training phase and facilitate a smooth transition to domain-specific tasks, we then design a Boost CPT phase. In this phase, we combine a very high-quality medical pre-training corpus with open-source medical SFT data at a 1:1 ratio, with a token-level distribution of 4:6 for Chinese:English. Notably, throughout these two phases, we progressively increase the proportion of Chinese data.

\subsection{Supervised Fine-Tuning}
To enhance model's ability to follow medical instructions and better adapt to specific medical scenarios, we conduct the SFT. This process involves using conversational-style data (comprising both queries and responses) to finetune the pretrained LLM. In the following sections, we will explore the details of data construction and training methods.

\subsubsection{Data Construction}
Our SFT dataset comprises a diverse array of question types, including multiple-choice questions from medical exams, single-turn disease diagnoses, and multi-turn health consultations. It integrates data from seven publicly available sources: Chinese Medical Dialogue Data\footnote{https://github.com/Toyhom/Chinese-medical-dialogue-data}, Huatuo26M \cite{li2023huatuo26mlargescalechinesemedical}, MedDialog \cite{zeng-etal-2020-meddialog}, ChatMed Consult Dataset \cite{tian2023chimedgpt}, ChatDoctor \cite{li2023chatdoctor}, CMB\footnote{https://github.com/FreedomIntelligence/CMB}, and MedQA \cite{MedQA}. We preserve portions of authentic doctor-patient conversations and augment the dataset by rewriting the remaining content. For these rewrites, we use real-world medical scenarios as prompts and generate responses via GPT-4. We believe this ensures the diversity of the SFT dataset, which can help the CareBot better adapt to different types of medical problems and patient situations, thereby improving its performance in a variety of scenarios.

As stated in \citet{zhou2023lima}, a relatively small, high-quality dataset can be sufficient for fine-tuning LLMs, our focus is on efficiently filtering "good data" from massive data to achieve competitive performance with a minimal amount of data. Like standard data cleaning processes, our approach begins by removing duplicates and eliminating data associated with security concerns such as violence, bias, and pornography. In the following sections, we specifically introduce the data selection methods.

\noindent\textbf{Single-turn Medical Dialogue Data} \quad Following \citet{liu2024what,zeng2024automatic}, we believe that "good data" should have a complex instruction and a high-quality response. Therefore, We adopt the approach from Deita \cite{liu2024what}, which employs a complexity model and a quality model to score each instance along two dimensions: instruction complexity and response quality. The complexity model assigns a complexity score $c_i$ to each instance, while the quality model assigns a quality score $q_i$, reflecting the quality of the response. By multiplying $c_i$ with $q_i$, we combine the complexity score and quality score to obtain a comprehensive score, that is, $s_i=c_i*q_i$. Finally, we set a score threshold to select the most effective data instances in the massive data pool.

\noindent\textbf{Multi-turn Medical Dialogue Data} \quad For multi-turn dialogues, we initially use Deita to compute the score $s_i$ for each individual turn and then average these scores to derive the final score for the entire dialogue. However, we identify two specific challenges in multi-turn dialogues compared to single-turn dialogues: (1) The low correlation between different turns can negatively affect the relevance of earlier information for subsequent turns; (2) Excessive correlation between turns can lead to significant context duplication and redundant information. To address these issues, we propose the ConFilter method, which uses a score $CF$ based on cross-entropy loss, to assess the influence of historical information on each turn. The details of this approach are outlined as follows:

In the instruction-tuning process, the loss of a sample pair $(H, T)$ is calculated by continuously predicting the next tokens in the current turn $T$ given their previous tokens and the history information $H$:

\begin{equation}
\begin{aligned}
L_{\theta}(t_{i}|H) = -\frac{1}{N}\sum_{j=1}^{N}logP(w_{i}^j|H,w_{i}^1,w_{i}^2,...,w_{i}^{j-1};\theta)
\end{aligned}
\end{equation}
where $H=\{t_1,t_2,...t_{i-1}\}$, $t_i$ is the current turn, $w_{i}^{j}$ is the $j$-th token in the $i$-th turn, and $N$ is the number of tokens of the current turn. We define $L_{\theta}(t_{i}|H)$ as the Conditioned Information Score, which measures the ability to generate the current turn under the guidance of corresponding historical information. 

To measure the ability of LLM to generate this turn alone, we also define a Direct Information Score:
\begin{equation}
\begin{aligned}
L_{\theta}(t_{i}) = -\frac{1}{N}\sum_{j=1}^{N}logP(w_{i}^j|w_{i}^1,w_{i}^2,...,w_{i}^{j-1};\theta)
\end{aligned}
\end{equation}
We believe that the higher Direct Information Score may indicate that the turn is more challenging or complex. Finally, we try to estimate $CF$ by calculating the ratio between $L_{\theta}(t_{i})$ and $L_{\theta}(t_{i}|H)$.
\begin{equation}
\begin{aligned}
{CF}_{\theta}(H,T)=\frac{L_{\theta}(t_{i}|H)}{L_{\theta}(t_{i})}
\end{aligned}
\end{equation}
Here, if $CF>1$, it means historical information has a negative impact on current turn, that is, the correlation between contexts is very low. If $CF<1$, it means historical information has a positive impact on current turn, that is, the correlation between contexts is high. However, too small $CF$ means that the context is highly repeated and the information is highly redundant. We also set a threshold to filter the data.

\subsection{RLHF}
We enhance the model's capabilities using Direct Preference Optimization (DPO) \cite{rafailov2023direct} after the SFT stage. To align the model's output with human preferences while preserving the foundational abilities gained during the CPT and SFT stages \cite{lu2024online}, we construct subjective preference data and objective preference data using samples that have the same distribution as the SFT dataset:

\noindent\textbf{Subjective Preference Data} \quad We aim to construct dpo pairs where the chosen response aligns closely with human preferences. For each prompt, we first ask GPT-4 to respond as a professional and helpful doctor. Then, using GPT-4, we evaluate the superiority or inferiority of the original response and this newly generated response from the prompt. The evaluation considers four aspects: fluency, relevance, completeness, and proficiency in medical. We select the superior response as the chosen response for the dpo pair and the inferior response as the rejection response.

\noindent\textbf{Objective Preference Data} \quad While RLHF can guide LLMs to align with human expectations, numerous studies show that this method can cause LLMs to forget abilities acquired during pre-training and SFT stages \cite{bai2022training, dong2023abilities}, leading to an "alignment tax" \cite{dong2023raft,sun2024principle}. To mitigate this issue, we construct objective preference data. Specifically, for objective prompts with known ground truth answers, we consider the ground truth as the chosen response and randomly select incorrect answers from the remaining options as rejection responses. For instance, in multiple-choice questions, if the ground truth is option A, we randomly select from options B, C, and D to construct the rejection response.

\begin{table*}[t]
\centering
\resizebox{0.8\textwidth}{!}{%
\begin{tabular}{l|cc|ccc|ccc|c}
\hline
\multirow{2}{*}{\textbf{Models}} & \multicolumn{2}{c|}{\textbf{English}} & \multicolumn{3}{c|}{\textbf{Chinese}}  \\
\cline{2-7} & \textbf{MedQA}  & \textbf{MMLU-Med} & \textbf{CMB} & \textbf{CMMLU-Med} & \textbf{C-Eval-Med} & \textbf{Avg.} \\ 
\hline
\hline
\textbf{ChatGPT} & 52.24 & 69.96 & 43.26 & 50.37 & 48.80 & 52.93 \\
\hline
\textbf{HuatuoGPT-7B} & 16.63 & 25.62 & 19.68 & 23.11 & 25.66 & 22.14\\
\textbf{Zhongjing-13B} & 11.28 & 16.90 & 20.39 & 23.85 & 30.09 & 20.50\\
\textbf{MedAlpaca-7B} & 49.74 & 62.72 & 23.29 & 25.38 & 27.43 & 37.71 \\
\textbf{Biomistral-7B} & 50.60$\dag$ & 59.08$\dag$ & 23.83 & 26.55 & 25.67 & 37.15\\
\textbf{HuatuoGPT ll-7B} & 41.13 & 51.44 & \textbf{60.39}  & \textbf{59.08} & 62.40 & 54.89 \\
\hline
\textbf{CareBot-Chat} & \textbf{63.71} & \textbf{71.53} & 52.50 & 56.42 & \textbf{63.72} & \textbf{61.58} \\
\textbf{CareBot-Chat (RL)} & 63.63 & 71.44 & 52.45  & 56.58 & 62.83 & 61.39 \\
\hline
\end{tabular}%
}
\caption{The results of five medical concept knowledge benchmarks. $\dag$ means the result of 3-shot (consistent with the original paper) and others are 0-shot. All scores are averaged over three random runs. (p$<$0.05 under t-test)}
\label{Table: main_results}
\end{table*}

\section{Experimental Setup}
\subsection{Baselines}
We conduct a comparative analysis of our model against most representative open-source medical LLMs including HuatuoGPT-7B \cite{zhang2023huatuogpttaminglanguagemodel}, Zhongjing-13B \cite{yang2023zhongjingenhancingchinesemedical}, MedAlpaca-7B \cite{han2023medalpacaopensourcecollection}, BioMistral-7B \cite{BioMistral}, and HuatuoGPT ll-7B \cite{chen2023huatuogptiionestagetrainingmedical}. These models are specifically designed for medical applications, showcasing robust open-domain chat capabilities and applicability to various medical scenarios. Additionally, we also compare results from the closed-source model GPT-3.5-turbo. More details can be found in Appendix E.
\subsection{Medical Benchmark}
We comprehensively evaluate CareBot's medical capabilities from two aspects, one is medical concept knowledge, and the other is medical consultation ability. For medical concept knowledge, CareBot is evaluated using three popular Chinese medical benchmarks (CMB \cite{CMB}, CMMLU-Med \cite{CMMLU}, C-Eval-Med \cite{ceval}) and four English medical benchmarks (MedQA \cite{MedQA}, MMLU-Med \cite{MMLU}, MedMCQA \cite{pal2022medmcqa}, and PubMedQA \cite{jin2019pubmedqa} test set). Accuracy is served as the primary evaluation metric for this aspect. For single-turn medical consultation questions, we use the Huatuo26M-test \cite{li2023huatuo26mlargescalechinesemedical}, evaluating responses via HuatuoEval \cite{chen2023huatuogptiionestagetrainingmedical} for pairwise comparisons.Additionally, multi-turn medical consultation questions are assessed using CMtMedQA \cite{yang2023zhongjingenhancingchinesemedical} and CMB-Clin \cite{CMB}. Consistent with \citet{CMB}, the model's responses are rated based on the fluency, relevance, completeness and medical proficiency of the reference answers. More details can be found in Appendix E.


\begin{figure}[htp]
 \centering
\resizebox{0.35\textwidth}{!}{
 \includegraphics{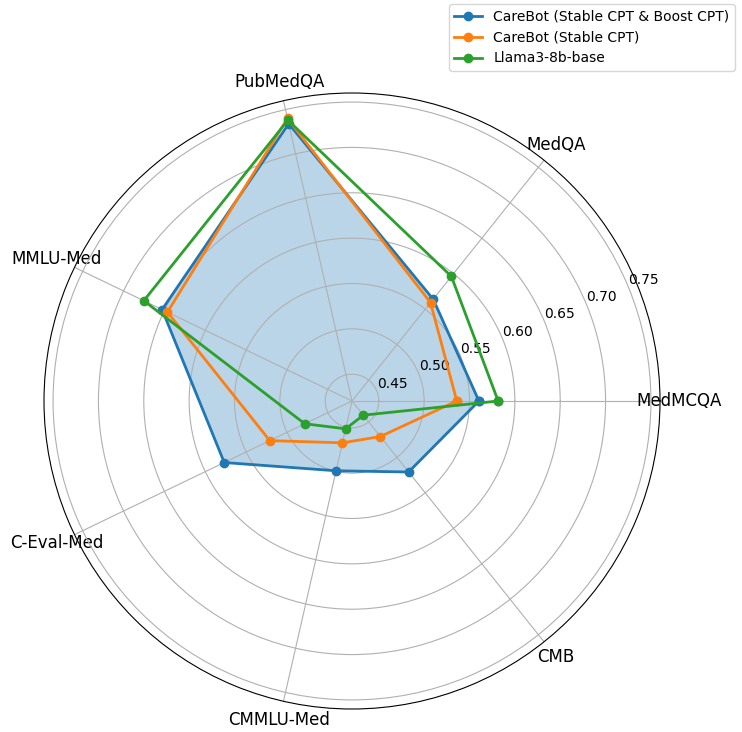}
 }
 \caption{The performance of seven benchmarks for our CPT model, CareBot. }
 \label{fig:CPT}
\end{figure}

\section{Experimental Results for CPT}
In Figure \ref{fig:CPT}, we evaluate our CPT model, CareBot, on seven common medical benchmarks. Considering that our goal is to train a medical model that performs well in both Chinese and English, we strive to improve Chinese medical ability while ensuring that English medical ability of the model is slightly reduced. We observe that for English benchmarks (MMLU-Med, PubMedQA, MedQA, MedMCQA), the performance of CareBot (Stable CPT) and CareBot (Stable CPT \& Boost CPT) shows a slight decrease. This is expected, given that the LLaMA-8B-base model already has strong English capabilities. However, for Chinese benchmarks (C-Eval-Med, CMMLU-Med, CMB), our models demonstrate significant improvements, with particularly notable gains in models trained using the two-stage approach. This confirms that our two-stage CPT strategy effectively integrates medical domain knowledge into the model, resulting in robust enhancements to its Chinese medical capabilities.

\begin{table*}[]
\centering
\resizebox{0.7\textwidth}{!}{%
\begin{tabular}{l|cccc|c}
\hline
\textbf{Models} & \textbf{Fluency} & \textbf{Relevance} & \textbf{Completeness} & \textbf{Proficiency} & \textbf{Avg.} \\
\hline
\textbf{ChatGPT} & 4.94 & 4.21 & 3.98 & 4.04 & 4.29\\
\hline
\textbf{HuatuoGPT-7B} & 4.94 & 3.28 & 3.22 & 4.19 & 3.91\\
\textbf{Zhongjing-13B} & 2.26 & 1.68 & 1.62 & 2.63 & 2.05\\
\textbf{MedAlpaca-7B} & 4.35 & 2.18 & 1.97 & 3.29 & 2.95\\
\textbf{Biomistral-7B} & 4.48 & 2.45 & 2.17 & 3.55 & 3.16\\
\textbf{HuatuoGPT ll-7B} & 4.96 & 3.41 & 3.47 & 4.27 & 4.03\\
\hline
\textbf{CareBot-Chat} & \textbf{4.99} & 4.67 & 4.20 & 4.26 & 4.53\\
\textbf{CareBot-Chat (RL)} & 4.96 & \textbf{4.70} & \textbf{4.31} & \textbf{4.34} & \textbf{4.58}  \\
\hline
\end{tabular}%
}
\caption{The scores of different models on CMtMedQA.}
\label{Table: CMtMedQA}
\end{table*}

\begin{table*}[]
\centering
\resizebox{0.7\textwidth}{!}{%
\begin{tabular}{l|cccc|c}
\hline
\textbf{Models} & \textbf{Fluency} & \textbf{Relevance} & \textbf{Completeness} & \textbf{Proficiency} & \textbf{Avg.} \\
\hline
\textbf{ChatGPT} & 4.78 & 3.75 &  3.77 & 4.01 & 4.08\\
\hline
\textbf{HuatuoGPT-7B} & 4.69 & 3.25 & 3.08 & 4.00 & 3.76\\
\textbf{Zhongjing-13B} & 3.14 & 2.14 & 1.89 & 3.11 & 2.57\\
\textbf{MedAlpaca-7B} & 3.31 & 1.69 & 1.49 & 2.58 & 2.27\\
\textbf{Biomistral-7B} & 3.79 & 2.10 & 1.82 & 3.09 & 2.70\\
\textbf{HuatuoGPT ll-7B} & 4.75 & 3.43 & 3.43 & 4.22 & 3.96\\
\hline
\textbf{CareBot-Chat} & \textbf{4.82} & 4.20 & 3.68 & 4.16 & 4.22  \\
\textbf{CareBot-Chat (RL)} & 4.74  & \textbf{4.28} & \textbf{4.01} & \textbf{4.21} & \textbf{4.31}  \\
\hline
\end{tabular}%
}
\caption{The scores of different models on CMB-Clin.}
\label{Table: CMB-Clin}
\end{table*}

\section{Experimental Results for Alignment}
\subsection{Results for Medical Concept Knowledge}
We present the results from five widely used benchmarks in Table \ref{Table: main_results}. For English benchmark MedQA, our model CareBot-Chat outperforms ChatGPT by 11.47 points and BioMistral, the strongest open-source medical LLM, by 13.11\%. For MMLU-Med, CareBot-Chat achieves a 1.57\% improvement over ChatGPT and surpasses MedAlpaca by 8.81 points. For English benchmarks CMB and CMMLU-Med, HuatuoGPT II emerges as the top-performing model and our model does not have an advantage over it. For C-Eval-Med, our CareBot only achieves competitive results with HuatuoGPT II. This is because CareBot is built on LLaMA3-8B, an LLM with inherent strengths in English. Therefore, its performance in Chinese aligns with expectations. Nevertheless, in terms of average scores, CareBot-Chat exceeds HuatuoGPT II, the leading open-source medical model, by 6.69\%, and ChatGPT by 8.65\%. These outcomes underscore CareBot's exceptional performance in medical applications, establishing it as a significant contributor to the field of medical AI.

\subsection{Results for Medical Consultation Ability}

\noindent\textbf{Multi-turn Dialogue}\quad In Table \ref{Table: CMtMedQA} and \ref{Table: CMB-Clin}, we present the results for the multi-turn dialogue benchmarks CMtMedQA and CMB-clin, respectively. Overall, the performance of two baselines is basically the same, with notable performance from HuatuoGPT and HuatuoGPT ll. Across four dimensions, fluency and proficiency scores are particularly high, indicating coherent responses and a solid grasp of medical terminology by the medical LLMs. However, relevance and completeness scores are lower, suggesting room for improvement in providing highly relevant and comprehensive answers tailored to specific questions. Our model, CareBot-Chat, achieves strong performance across all dimensions, averaging scores of 4.35 and 4.22 respectively, with notable strengths in relevance and completeness. This underscores the effectiveness of our high-quality SFT dataset and our proposed multi-turn dialogue selection method, significantly enhancing the model's contextual understanding and ensuring dialogue coherence and consistency. We further analyze the advantages of our models in multi-turn dialogues in Section Advantages of Multi-turn Dialogue.

\begin{figure}[t]
\centering
\subcaptionbox{CareBot-Chat\label{CareBot-Chat}}{\includegraphics[width =0.9\linewidth]{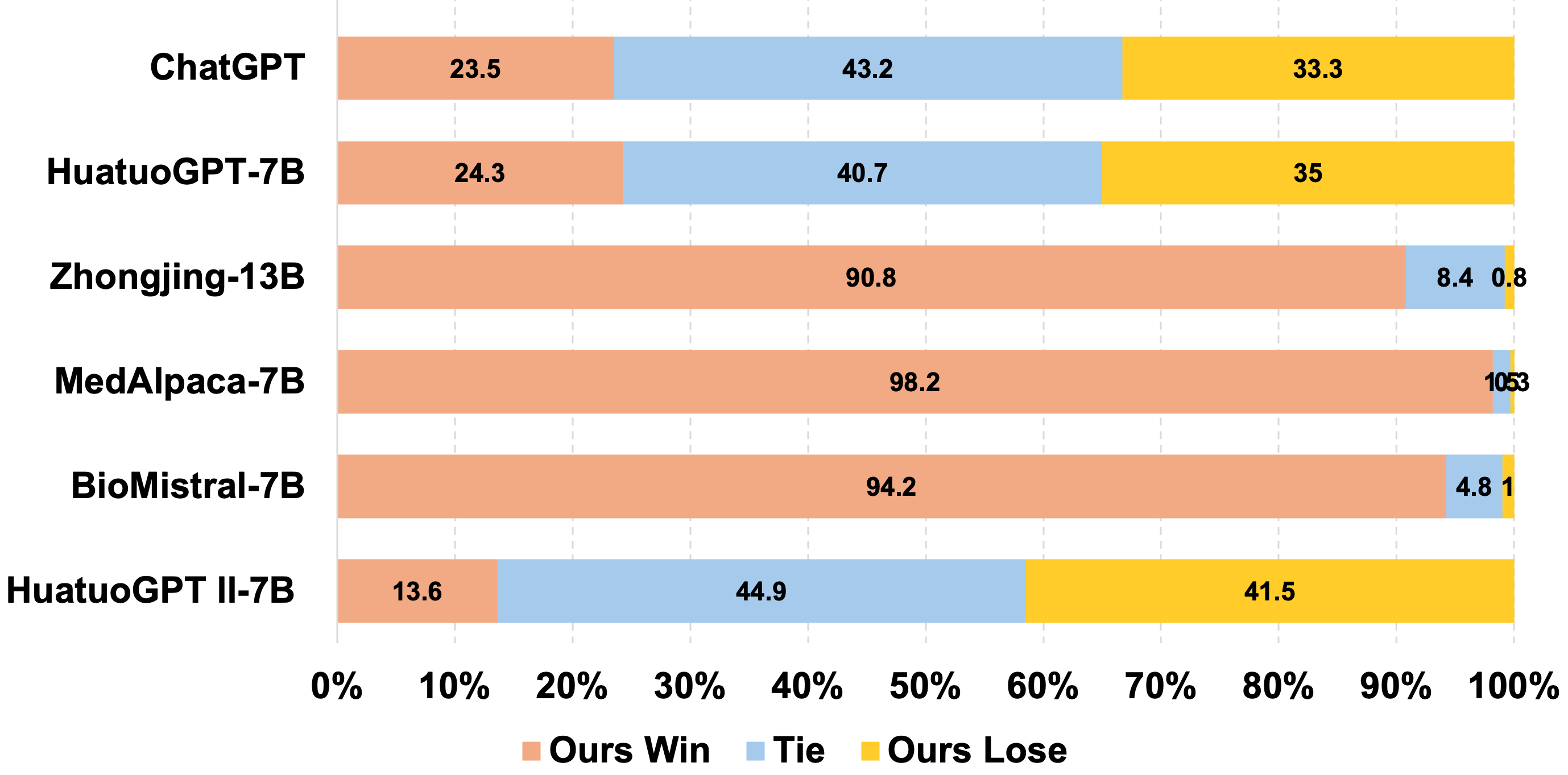}}

\subcaptionbox{CareBot-Chat (RL)\label{CareBot-Chat (RL)}}{\includegraphics[width =0.9\linewidth]{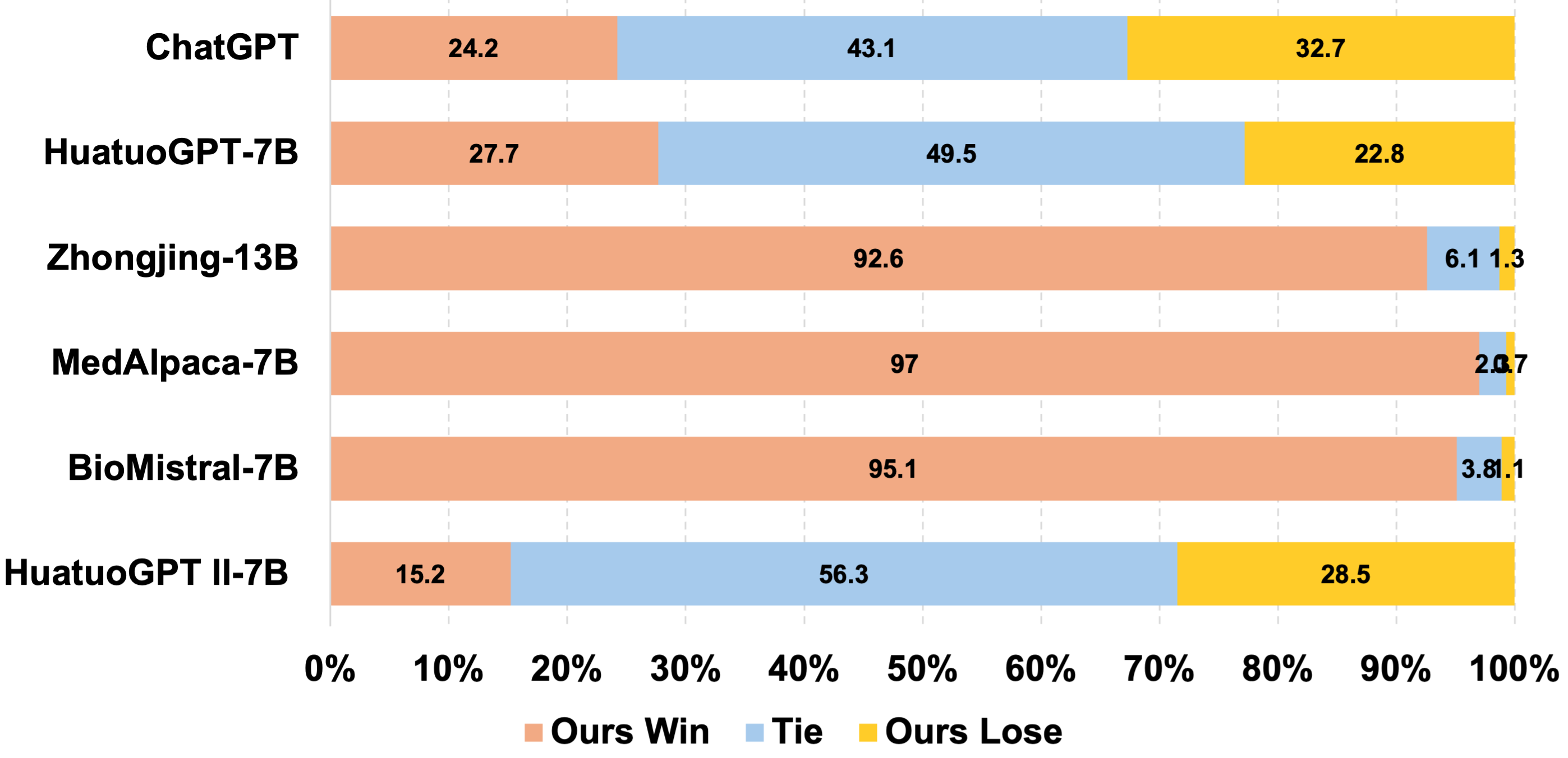}}
\caption{Comparison of CareBot-Chat and CareBot-Chat (RL)'s predicted answers and other baselines' predicted answers on single-turn dialogues from Huatuo26M-test.}
\label{fig:Chat_pair}
\end{figure}



\noindent\textbf{Single-turn Dialogue}\quad Figure \ref{fig:Chat_pair} (a) displays the comparison results between our CareBot-Chat and various baselines on the single-turn dialogue benchmark Huatuo26M-test. It is evident that CareBot-Chat achieves comparable performance to ChatGPT and HuatuoGPT, indicating that the baseline HuatuoGPT performs similarly to ChatGPT under this evaluation framework. Moreover, our model significantly outperforms Zhongjing, MedAlpaca, and BioMistral. The lower performance of MedAlpaca and BioMistral can be attributed to their limited Chinese language capabilities and inadequate medical SFT data. It is noteworthy that Zhongjing, incorporating pre-training, SFT, and RLHF stages, performs poorly. This is likely due to the two training stages of the base model, leading to catastrophic forgetting or model degradation. Additionally, its instruction fine-tuning data is both limited and of inferior quality, which explains why a model with larger parameters than our CareBot performs substantially worse. However, we observe that our model only competes with HuatuoGPT ll-7B in the HuatuoEval framework. We will further discuss this phenomenon in Section Case Study. In Figure \ref{fig:Chat_pair} (b), we also compare our CareBot-Chat (RL) model with other baselines, and basically achieve the same performance as CareBot-Chat.

\section{Analysis and Discussion}

\subsection{One Stage CPT vs Two Stage CPT}
\begin{figure}[htp]
 \centering
\resizebox{0.35\textwidth}{!}{
 \includegraphics{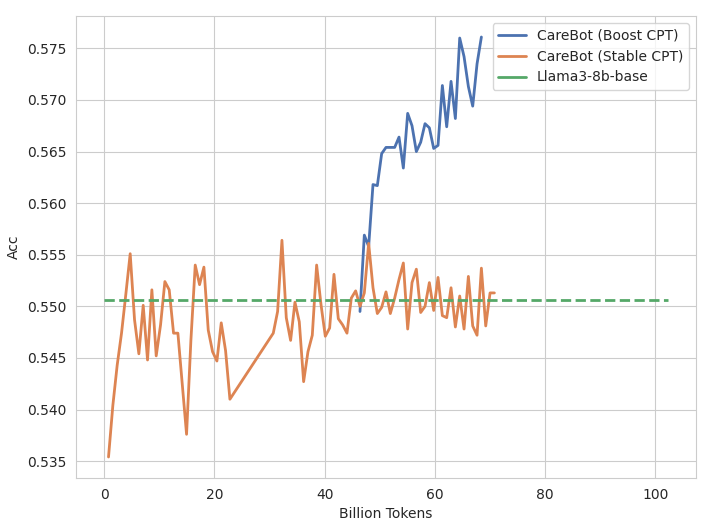}
 }
 \caption{Comparison of the loss between our proposed two-stage CPT and the plain CPT.}
 \label{fig:two_stage_CPT}
\end{figure}

Figure \ref{fig:two_stage_CPT} illustrates the changes in Acc for the standard CPT (represented by the orange line) and our two-stage CPT method (shown with both the orange and blue lines). Initially, as the number of training tokens increases, the Acc rises but fluctuates significantly. After reaching 45B tokens, the Acc stabilizes around the LLaMA-8B-base level, indicating that the medical CPT has reached a relatively stable state. Introducing the Boost CPT stage results in a marked and consistent improvement in Acc. In contrast, continuing training with 20B additional tokens using the Stable CPT approach shows minimal changes in performance. These findings strongly demonstrate that our two-stage CPT method effectively facilitates a smooth transition from general knowledge to domain-specific knowledge, from English to Chinese, and from PT to SFT.

\subsection{Advantages of Multi-turn Dialogue}
\label{sec:Advantages of Multi-turn Dialogue}
To investigate the reasons behind our model's strong performance in multi-turn dialogues, we analyze the quality of each turn of responses, as shown in Table \ref{Table: each_turn} (Appendix G). Our analysis reveals that, except for the first turn, CareBot-Chat's average score was 0.25 lower than that of HuatuoGPT II. However, in subsequent turns, CareBot-Chat consistently outperforms HuatuoGPT II. Furthermore, our advantage becomes increasingly pronounced as the number of dialogue turns increases. Specifically, in terms of fluency and proficiency, our model performs comparably to HuatuoGPT II. In contrast, HuatuoGPT II's performance in relevance and completeness significantly deteriorates with more turns, while CareBot-Chat maintains stable performance. This highlights CareBot-Chat’s superior capability in multi-turn dialogues, demonstrating its better understanding of dialogue context. We attribute this advantage to our carefully curated multi-turn dialogue SFT data, which retains dialogues with contextual relevance without excessive repetition. Additionally, Figure \ref{fig:CareBot_multi_our} and \ref{fig:CareBot_multi_Huatuo_II} in Appendix G provide comparisons of responses from CareBot-Chat and HuatuoGPT II in the same multi-turn dialogue, further confirming our model's effectiveness. Notably, in the first turn, HuatuoGPT II's response is longer and more detailed than ours. However, from the second turn, while HuatuoGPT II's answers remain detailed, they become increasingly irrelevant and lack coherence with the ongoing dialogue. Generally, our results strongly indicate that CareBot-Chat has a significant advantage in multi-turn dialogues, demonstrating superior contextual understanding and coherence.

\begin{figure}[t]
 \centering
\resizebox{0.35\textwidth}{!}{
 \includegraphics{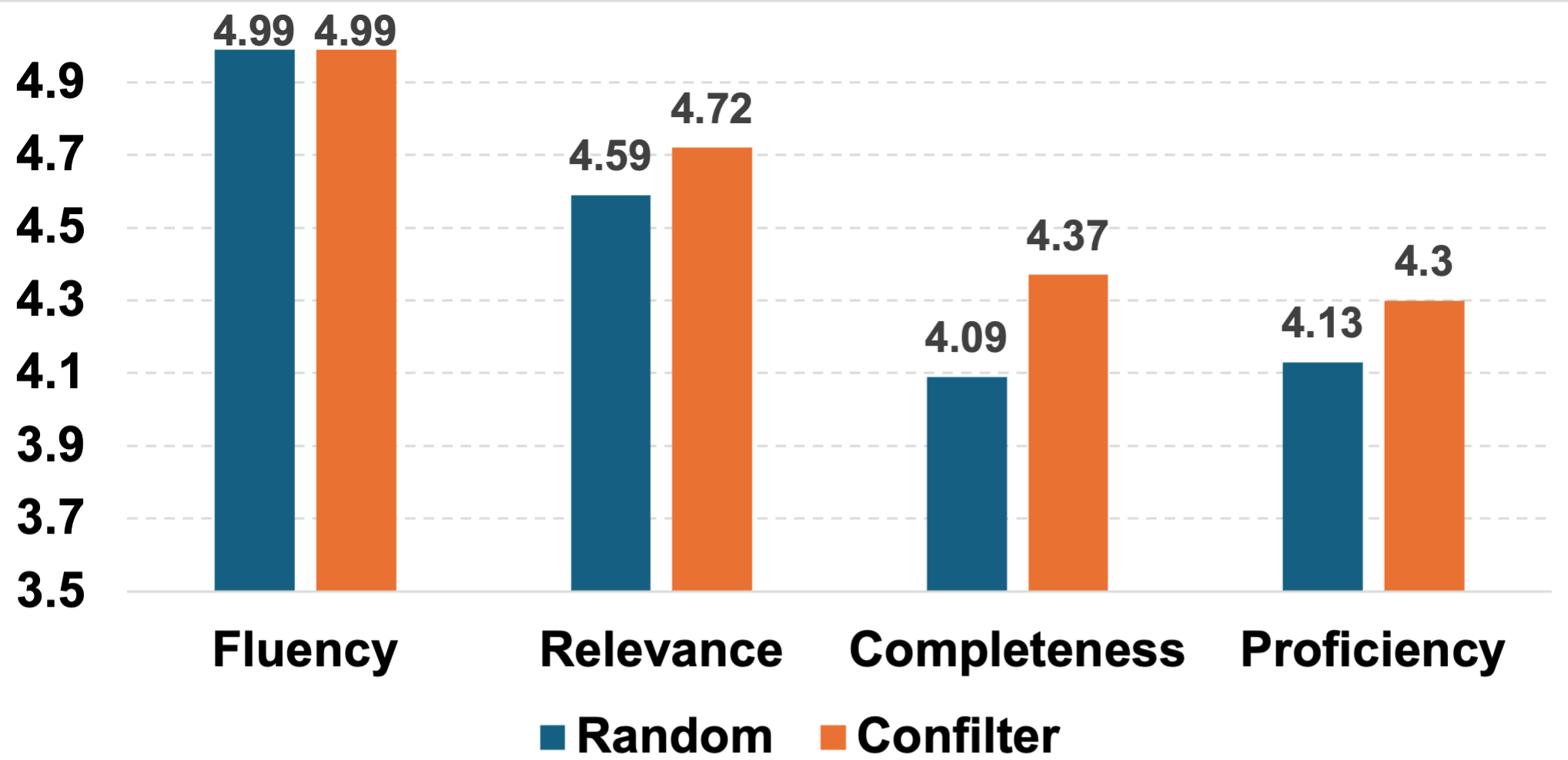}
 }
 \caption{The comparison of the high-quality multi-turn dialogues selected by our ConFilter and the randomly selected multi-turn dialogues. These are all from CMtMedQA.}
 \label{fig:high_low_pair}
\end{figure}

\subsection{Effect on ConFilter}
To evaluate the effectiveness of our multi-turn SFT data selection method, ConFilter, we conduct a comparative experiment. We finetune the pre-trained CareBot using two datasets: 110k high-quality multi-turn dialogues selected by ConFilter and 110k randomly selected samples from the original CMtMedQA dataset. As shown in Figure \ref{fig:high_low_pair}, both datasets achieve comparable fluency scores. However, in the other three dimensions, fine-tuning with the high-quality data significantly outperforms the randomly selected data. These results underscore the effectiveness of ConFilter in enhancing dialogue coherence and relevance by focusing on contextual correlations, thereby helping the model better understand and maintain conversation history.

\subsection{Case Study}
Figure \ref{fig:Chat_pair} (a) illustrates our Carebot does not hold a significant advantage over HuatuoGPT II. Here, we present a case study. As depicted in Figure \ref{fig:case study} (Appendix H), we provide an example where the output of CareBot-Chat is deemed inferior to that of HuatuoGPT II under the HuatuoEval evaluation framework. However, upon closer examination, we found there are some issues in HuatuoGPT II's response, particularly in the statement "such as 100\% skim milk or low-fat milk. These milks usually contain more protein and calcium, and have lower sugar and fat content." Firstly, this statement itself is an hallucination, that is, skim milk or low-fat milk does not contain more protein and calcium, but only has lower fat content. Furthermore, medical professionals confirm that whole milk is more suitable for infants due to its nutritional benefits, including fats crucial for development. This aligns with responses generated by GPT-4. Additionally, HuatuoGPT II often includes repetitive content such as "It is better to choose milk designed specifically for children," which, despite mimicking a doctor's tone and offering longer responses, sometimes lacks relevance and completeness. This approach occasionally introduces ambiguities. 
\section{Conclusion}
In this paper, we propose CareBot, a bilingual medical LLM, which is designed to enhance medical diagnostics, treatment planning, and medical education. To bridge gaps between data with different distributions, we design a novel two-stage continuous pre-training (CPT) approach, i.e., stable CPT and boost CPT. A model for evaluating data quality during CPT, DataRater, is also proposed. Besides, we further present ConFilter for selecting high-quality multi-turn dialogues, which is crucial to improving the model's ability to handle more complex dialogues. CareBot’s performance, validated through extensive testing on Chinese and English medical benchmarks, demonstrates significant improvements in medical consultation and teaching, showcasing the effectiveness of its training strategies and datasets.

\section*{Acknowledgments}
We thank all anonymous reviewers. This work was supported by National Science and Technology Major Project No.2022ZD0116314.

\bibliography{aaai25}
\clearpage

\section{Appendix}

\subsection{A. Domain Classification}
\label{sec: Domain Classification}
\begin{figure}[htp]
 \centering
\resizebox{0.45\textwidth}{!}{
 \includegraphics{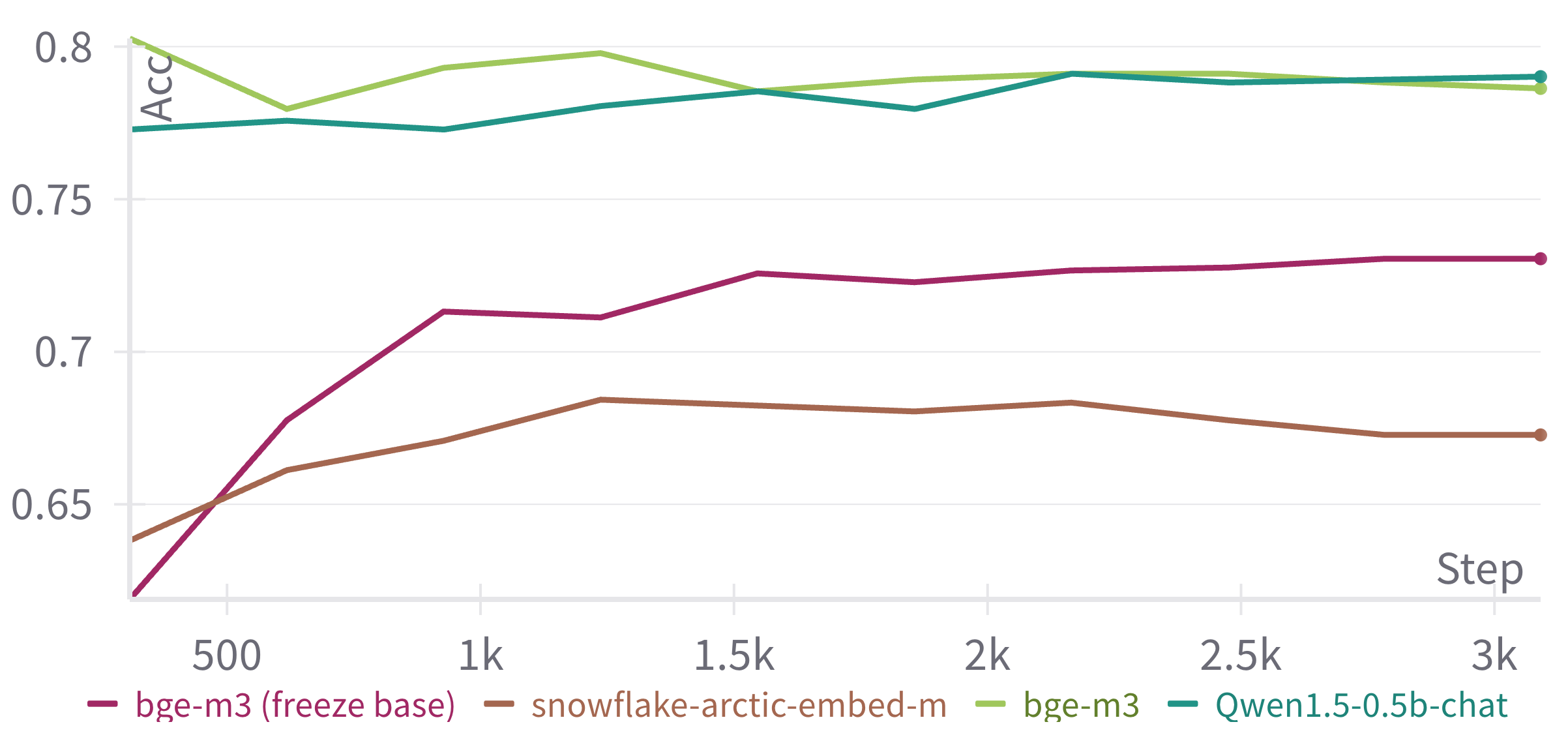}
 }
 \caption{Domain classification models for CPT data collection and decontamination.}
 \label{fig:domain-classification}
\end{figure}

In this section, we train a domain classifier on different pre-trained models, as shown in Figure\ref{fig:domain-classification}. Finally, we choose bge-m3 with the best Acc performance as the final choice, with a batch size of 64, lr of 1e-5 and the training epoch of 10. The validation accuracy is 86\%.

\subsection{B. LLM-based Data Quality Filtering}
\begin{figure}[htp]
 \centering
\resizebox{0.45\textwidth}{!}{
 \includegraphics{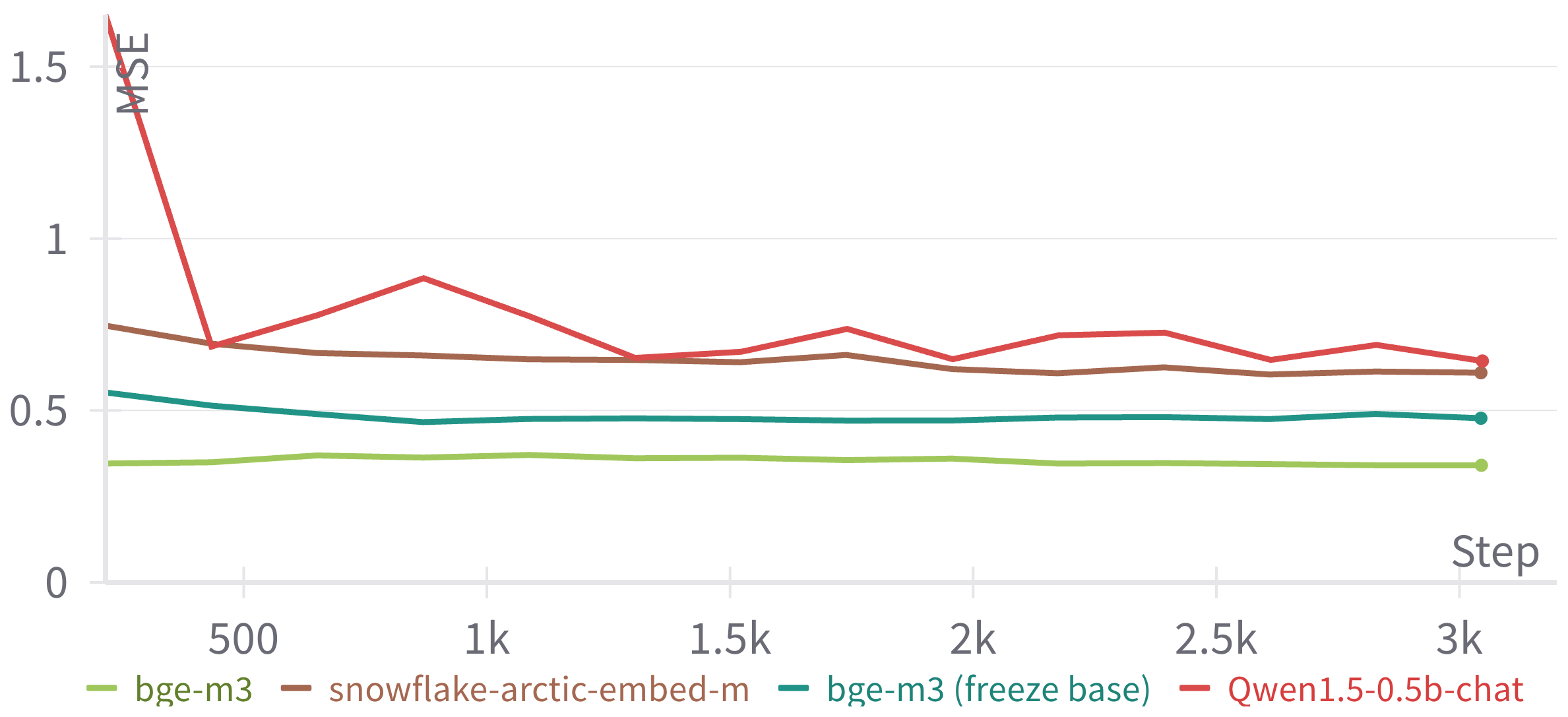}
 }
 \caption{Data quality filtering models for CPT data collection and decontamination.}
 \label{fig:data-quality-filter}
\end{figure}

In this section, we train a data quality assessment model,DataRater, on different pre-trained models, as shown in Figure\ref{fig:data-quality-filter}. Finally, we also choose bge-m3 with the lowest MSE as the final choice, with a batch size of 64, lr of 1e-5 and the training epoch of 10.

\subsection{C. Data Ratio Search in CPT Stage}
\begin{figure}[htp]
 \centering
\resizebox{0.45\textwidth}{!}{
 \includegraphics{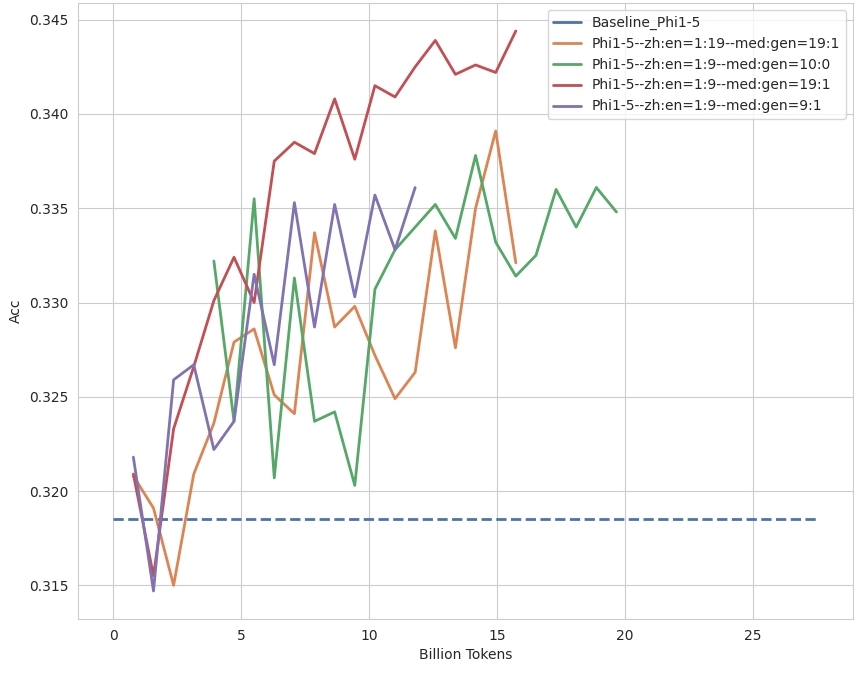}
 }
 \caption{Data Ratio Search for Chinese and English, and domain-specific data and general data in Stable CPT stage.}
 \label{fig:Data Ratio Search}
\end{figure}

In this section, we explore the data ratio on the small-size model Phi1.5 in Figure \ref{fig:Data Ratio Search}. We first keep the Chinese-English ratio at 1:9 and adjust the ratio of medical data and general data. We found that the best ratio is 19:1 for medical data to general data. Then, while keeping the ratio of medical data to general data at 19:1, we reduce the ratio of Chinese: English to 1:19. We found that this performance is not good. Therefore, we choose a Chinese:English ratio of 1:9 and a medical data:general data ratio of 19:1 as the final ratio.

\subsection{D. Training Details}
In this section, we give the training details of different stages as follows:

\noindent\textbf{CPT}\quad For stable CPT, we train on 3*8 NVIDIA A100-40G GPUs, using a batch-size of 4M, a learning rate of 8e-6, a maximum length of 4096, a cosine learning rate scheduler, a warmup-ratio of 0.1, and train for 18,000 steps.  For boost CPT, keeping other settings unchanged, we reduce the learning rate to 2e-6, and train for 7,000 steps. 

\noindent\textbf{SFT}\quad Our Chat model is based on CareBot and the training process has the following hyperparameters: sequence length set to 2048, batch size set to 128, and peak learning rate set to 2e-6 with cosine learning rate scheduler. To prevent overfitting, weight decay of 0.1 is applied and dropout is set to 0.1. Training is parallelized on 8 A100-40G NVIDIA GPUs using the AdamW optimizer with bf16 precision and ZeRO-3. We reserve 10\% of the training set for validation and get the best checkpoint after 2 epochs.

\noindent\textbf{DPO}\quad We construct a dataset of 12,727 DPO preference pairs, consisting of 9,019 subjective and 3,708 objective data samples. We train the model over two epochs using 8 NVIDIA Tesla A100 GPUs. The settings include a learning rate of 1e-7, a batch size of 96, and a beta of 0.03. Additionally, we employ a learning rate warmup and a cosine learning rate scheduler for optimization.

\begin{figure*}[t]
\centering
\subcaptionbox{CMB\label{CMB}}{\includegraphics[width =0.3\linewidth]{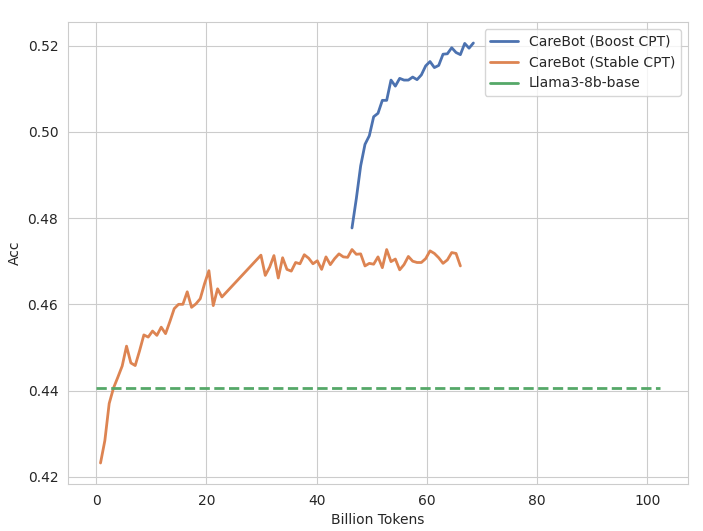}}\hfill
\subcaptionbox{CMMLU-Med\label{CMMLU-Med}}{\includegraphics[width =0.3\linewidth]{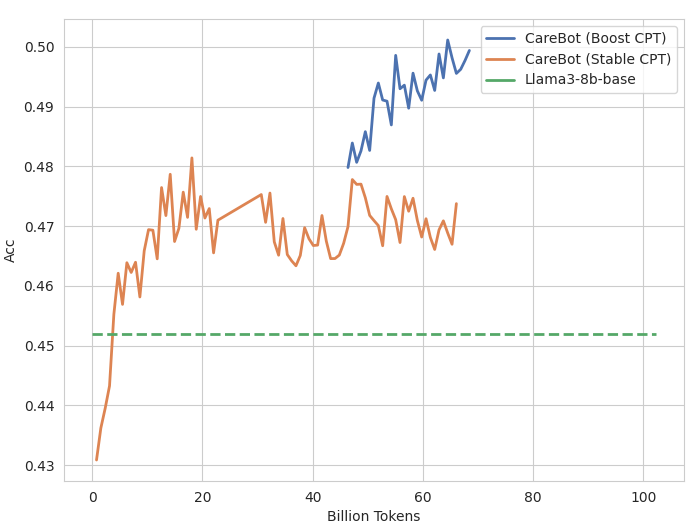}}\hfill
\subcaptionbox{CEval-Med\label{CEval-Med}}{\includegraphics[width =0.3\linewidth]{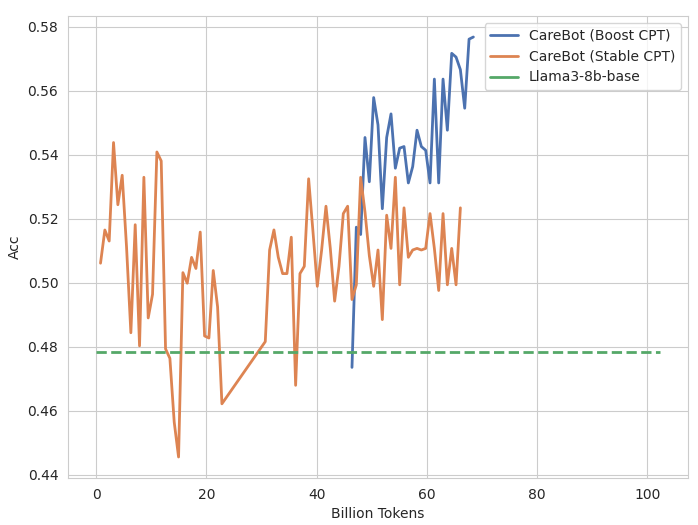}}

\subcaptionbox{PubMedQA\label{PubMedQA}}{\includegraphics[width =0.3\linewidth]{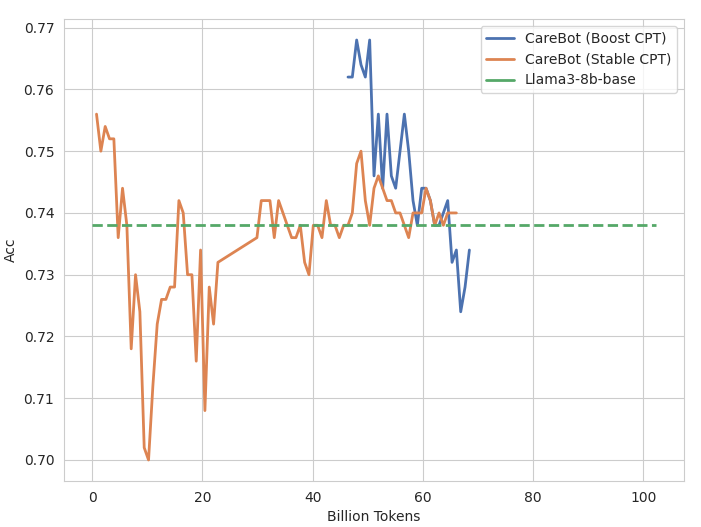}}\hfill
\subcaptionbox{MMLU-Med\label{MMLU-Med}}{\includegraphics[width =0.3\linewidth]{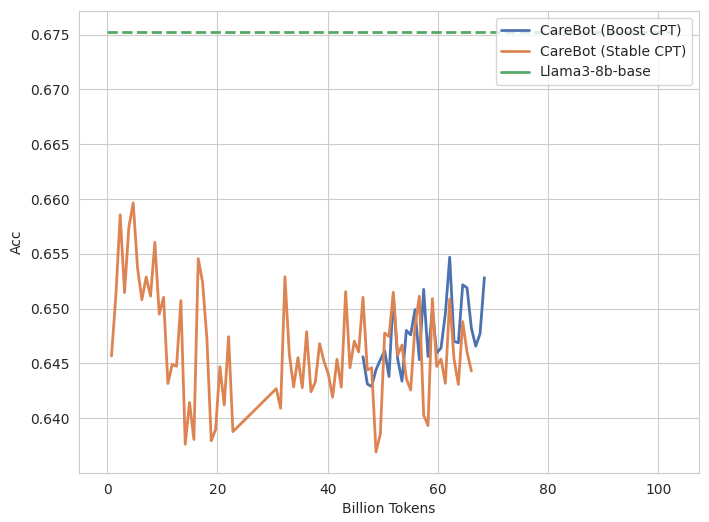}}\hfill
\subcaptionbox{MedQA\label{MedQA}}{\includegraphics[width =0.3\linewidth]{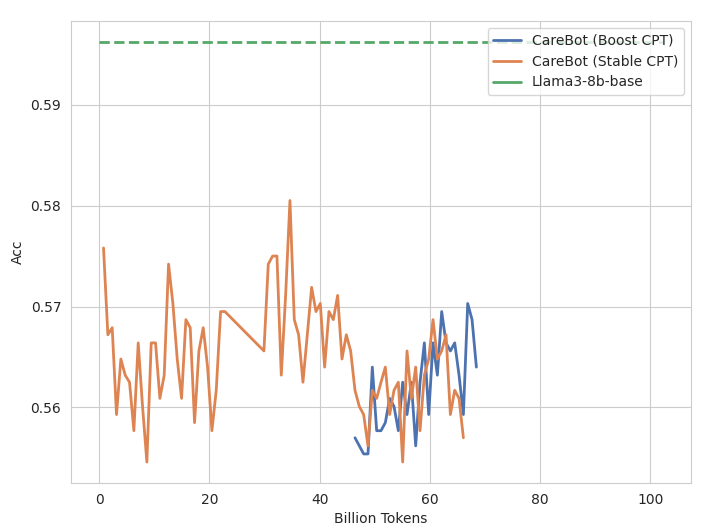}}

\subcaptionbox{Medmcqa\label{Medmcqa}}{\includegraphics[width =0.3\linewidth]{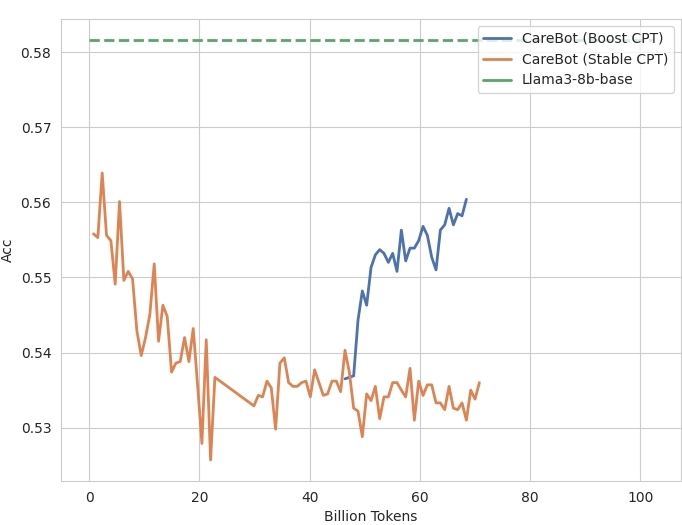}}\hfill

\hfill

\caption{Performance of two different CPT strategies on seven benchmarks.}
\label{Figure: two stage CPT_7}
\end{figure*}

\begin{table*}[t]
\centering
\resizebox{0.8\textwidth}{!}{%
\begin{tabular}{l|c|ccccc}
\hline
\multicolumn{2}{l|}{\textbf{Model}} & \textbf{Fluency} & \textbf{Relevance} &\textbf{Completeness} &\textbf{Proficiency} &\textbf{Avg.}  \\
\hline
\multirow{2}{*}{\textbf{Turn 1}} & HuatuoGPT ll & 5.00 & 4.78 & 4.75 & 4.54 & 4.77\\
& CareBot-Chat & 4.98 & 4.63 & 4.25 & 4.21 & 4.52 (-0.25) \\
\hline
\multirow{2}{*}{\textbf{Turn 2}} & HuatuoGPT ll & 4.97 & 3.74 & 3.72 & 4.26 & 4.17\\
& CareBot-Chat & 4.98 & 4.61 & 4.09 & 4.19 & 4.47 (+0.30)\\
\hline
\multirow{2}{*}{\textbf{Turn 3}} & HuatuoGPT ll & 4.93 & 2.95 & 2.96 & 4.09 & 3.73\\
& CareBot-Chat & 4.99 & 4.65 & 4.11 & 4.23 & 4.50 (+0.77)\\
\hline
\multirow{2}{*}{\textbf{Turn 4}} & HuatuoGPT ll & 4.94 & 2.69 & 2.85 & 4.17 & 3.66\\
& CareBot-Chat & 4.99 & 4.71 & 4.22 & 4.31 & 4.63 (+0.97)\\
\hline
\multirow{2}{*}{\textbf{Turn 5}} & HuatuoGPT ll & 4.92 & 2.54 & 2.83 & 4.23 & 3.63\\
& CareBot-Chat & 4.99 & 4.78 & 4.35 & 4.38 & 4.63 (+1.00)\\
\hline
\multirow{2}{*}{\textbf{Turn 6}} & HuatuoGPT ll & 4.94 & 2.55 & 2.83 & 4.29 & 3.65\\
& CareBot-Chat & 5.00 & 4.80 & 4.44 & 4.46 & 4.68 (+1.03)\\
\hline
\multirow{2}{*}{\textbf{Turn 7}} & HuatuoGPT ll & 4.93 & 2.93 & 3.07 & 4.24 & 3.79\\
& CareBot-Chat & 5.00 & 4.76 & 4.48 & 4.36 & 4.65 (+0.86)\\
\hline
\multirow{2}{*}{\textbf{Turn 8}} & HuatuoGPT ll & 4.92 & 2.54 & 2.62 & 4.08 & 3.54\\
& CareBot-Chat & 5.00 & 4.85 & 4.69 & 4.62 & 4.79 (+1.25)\\
\hline
\multirow{2}{*}{\textbf{Turn 9}} & HuatuoGPT ll & 4.80 & 2.80 & 3.00 & 3.80 & 3.63\\
& CareBot-Chat & 5.00 & 5.00 & 4.40 & 4.80 & 4.80 (+1.17) \\
\hline
\multirow{2}{*}{\textbf{Turn 10}} & HuatuoGPT ll & 4.96 & 3.41 & 3.47 & 4.27 & 3.18\\
& CareBot-Chat & 5.00 & 5.00 & 5.00 & 5.00 &5.00 (+1.82)\\
\hline
\end{tabular}%
}
\caption{The scores of each turn of responses on the CMtMedQA benchmark.}
\label{Table: each_turn}
\end{table*}

\subsection{E. Baselines and Benchmarks}
We use the following open-source medical large language models as baselines:

\textbf{HuatuoGPT-7B } is trained using real-world data and distilled data from ChatGPT, adopting RLMF (a method combining ChatGPT and doctor preferences) to leverage the advantages of mixed.

\textbf{Zhongjing-13B} is the first Chinese medical LLaMA-based LLM that implements an entire training pipeline from continuous pre-training, SFT, to Reinforcement Learning from Human Feedback (RLHF).

\textbf{MedAlpaca-7B} is a compilation of language models specifically fine-tuned for biomedical tasks. 

\textbf{BioMistral-7B} is a specialized LLM tailored for the biomedical domain, derived from Mistral 7B Instruct v0.1 and further pre-trained on PubMed Central.

\textbf{HuatuoGPT ll-7B} is a Chinese medical language model trained by a one-stage domain adaptation method.  

We use the following benchmarks for medical concept knowledge:

\textbf{CMB} is a collection of multiple-choice questions in Chinese, sourced from various professional mdedical qualification examinations. It encompasses questions from exams for physicians, nurses, technicians, pharmacists, undergraduate medical programs, and graduate entrance examinations. We utilize 11,200 questions from the test set to conduct a comprehensive, multi-level assessment of the model's medical knowledge.

\textbf{CMMLU-Med} is a comprehensive Chinese benchmark, from which we extract medical-related tasks to evaluate the model's performance. These tasks encompass various medical domains, including anatomy, clinical knowledge, college biology, college medicine, medical genetics, and professional medicine.

\textbf{C-Eval-Med} is a chinese multiple-choice dataset. We extracted tasks related to medicine from the validation set, such as basic medicine, clinical medicine, medical practice, and veterinary medicine to test the model's performance.

\textbf{MMLU-Med} is the english multi-subject multiple-choice dataset, from which we extract medical-related tasks to evaluate the model's performance. These tasks encompass various medical domains, including anatomy, clinical knowledge, college biology, college medicine, medical genetics, and professional medicine.

\textbf{MedQA} is a multiple-choice question dataset from the United States Medical Licensing Examination (USMLE). Its test set consists of 1,273 questions, which are used to assess a model's medical knowledge and reasoning skills required to obtain a medical license in the United States.

\textbf{MedMCQA} is a large-scale multiple-choice question and answer dataset, sourced from India's medical entrance exams (AIIMS/NEET). Its test set comprises 6,100 questions, enabling the evaluation of a model's general medical knowledge and reasoning abilities.

\textbf{PubMedQA} is a closed-domain question and answer dataset, where each question can be answered by referring to the relevant context from PubMed abstracts. We use 500 test questions from this dataset to evaluate a model's ability to understand and reason about biomedical literature.

We use the following benchmarks for medical consultation ability:

\textbf{Huatuo26M-test} is a large-scale Chinese Medical Question Answering (QA) dataset, and we use its test set to evaluate the model's capability in single-turn dialogues.

\textbf{CMtMedQA} is a large-scale dataset consisting of multi-turn medical dialogues in Chinese. To evaluate the model's ability to engage in complex dialogues and initiate proactive inquiries, we utilized approximately 1,000 samples from the dataset's test set.

\textbf{CMB-Clin} consists of 74 expertly curated medical case consultations derived from clinical diagnostic teaching materials. It evaluates the model's mastery and reasoning abilities in applying medical knowledge through multi-turn diagnostic dialogues.

\subsection{F. One Stage vs two Stage}
In this section, we present the performance of two CPT strategies on seven common benchmarks in Figure \ref{Figure: two stage CPT_7}.

\subsection{G. Analysis for Multi-turn Dialogue}
\begin{figure*}[htp]
 \centering
\resizebox{0.9\textwidth}{!}{
 \includegraphics{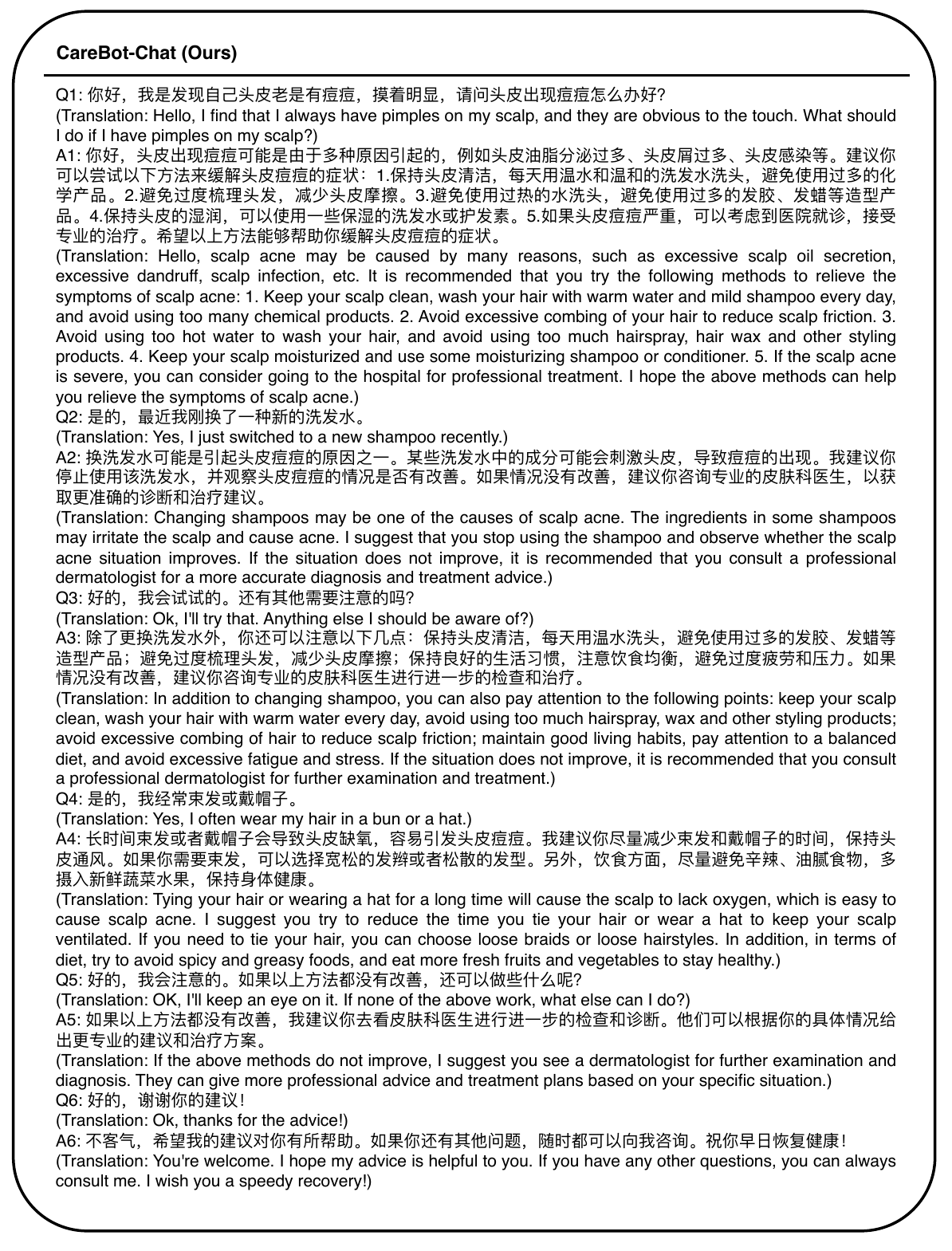}
 }
 \caption{An example of a multi-turn dialogue (CareBot-Chat).}
 \label{fig:CareBot_multi_our}
\end{figure*}

\begin{figure*}[htp]
 \centering
\resizebox{0.9\textwidth}{!}{
 \includegraphics{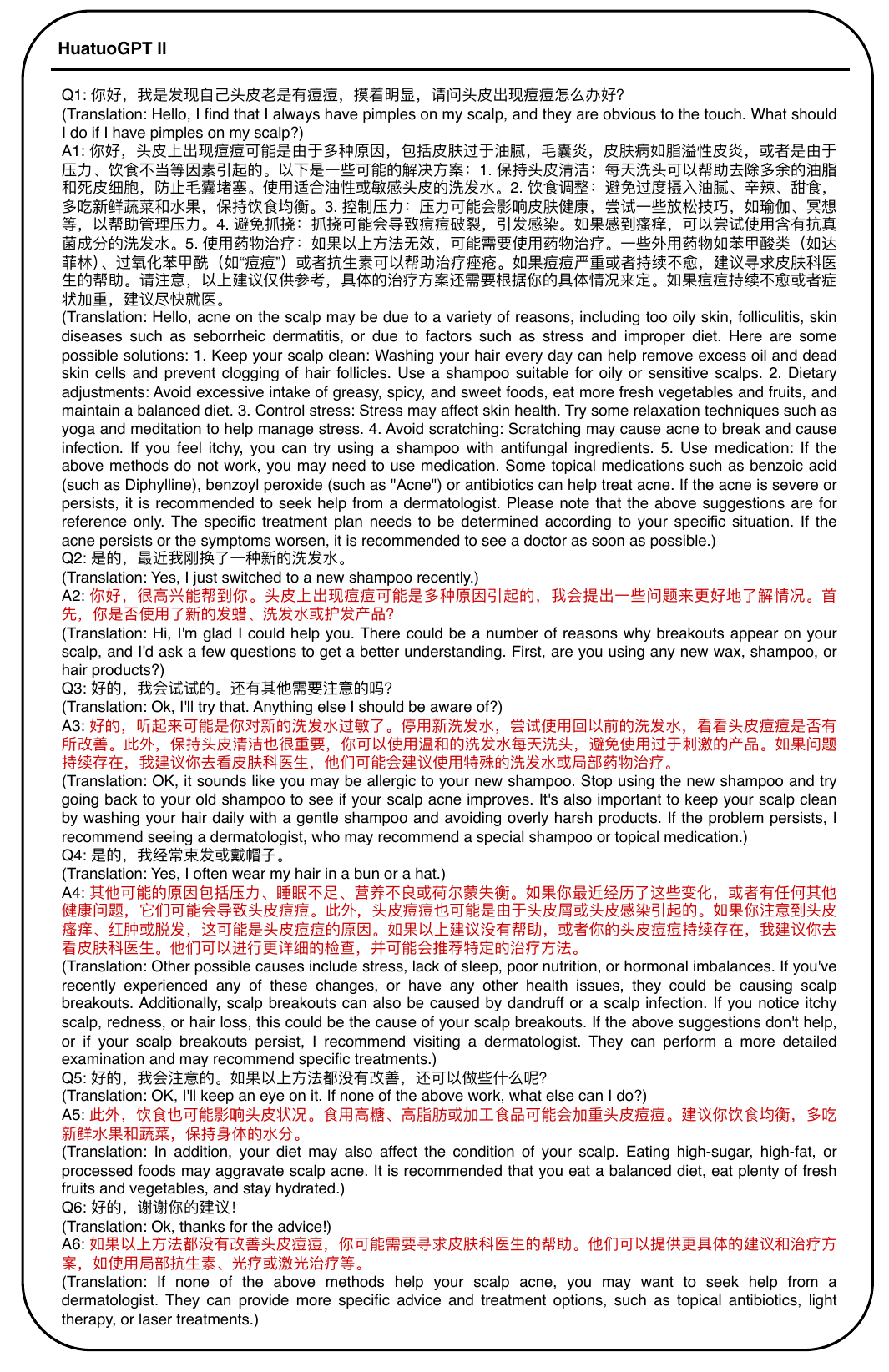}
 }
 \caption{An example of a multi-turn dialogue (HuatuoGPT ll).}
 \label{fig:CareBot_multi_Huatuo_II}
\end{figure*}

In this section, we first present the different responses of our CareBot-Chat and HuatuoGPT ll for the same dialogue, as shown in Figure \ref{fig:CareBot_multi_our} and Figure \ref{fig:CareBot_multi_Huatuo_II}. Then, in Table \ref{Table: each_turn}, we give the scores of each turn of responses on the CMtMedQA benchmark.

\subsection{H. Case Study for Single-turn Dialogue}
\begin{figure*}[htp]
 \centering
\resizebox{0.9\textwidth}{!}{
 \includegraphics{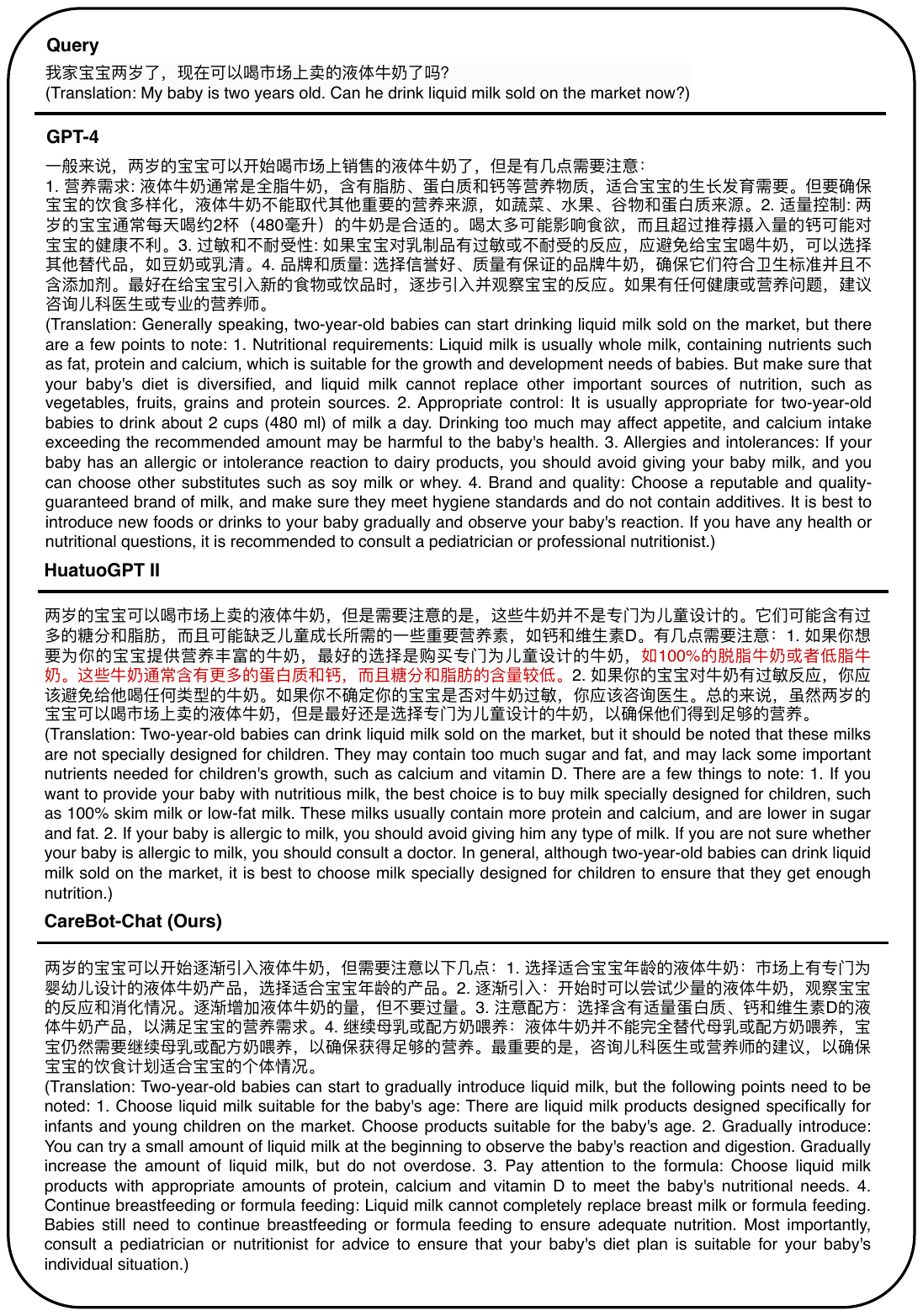}
 }
 \caption{An example of a multi-turn dialogue (HuatuoGPT ll).}
 \label{fig:case study}
\end{figure*}

In this section, we give an example of different models giving responses to the same single-turn dialogue in Figure \ref{fig:case study}.

\end{document}